\documentclass[twoside,leqno,twocolumn]{article}
\usepackage{ltexpprt}
\usepackage[T1]{fontenc}
\usepackage{graphicx, float}
\usepackage{xcolor, soul}
\usepackage{amssymb}
\usepackage{amsmath}
\usepackage{extarrows}
\usepackage{multirow}
\usepackage{array}
\usepackage{textcomp}

\begin{document}

\title{\Large Semantic Relation Classification via Bidirectional LSTM Networks with Entity-aware Attention using Latent Entity Typing }
\author{Joohong Lee\thanks{Department of Computer Science and Engineering, Hanyang University, Seoul, Republic of Korea, \{roomylee, ssw1591, cys\}@hanyang.ac.kr} \\
\and
Sangwoo Seo\footnotemark[1] \\
\and
Yong Suk Choi\footnotemark[1] \thanks{Corresponding author.}}
\date{}

\maketitle







\begin{abstract} \small\baselineskip=9pt 
Classifying semantic relations between entity pairs in sentences is an important task in Natural Language Processing (NLP).
Most previous models for relation classification rely on the high-level lexical and syntatic features obtained by NLP tools such as WordNet, dependency parser, part-of-speech (POS) tagger, and named entity recognizers (NER).
In addition, state-of-the-art neural models based on attention mechanisms do not fully utilize information of entity that may be the most crucial features for relation classification.
To address these issues, we propose a novel end-to-end recurrent neural model which incorporates an entity-aware attention mechanism with a latent entity typing (LET) method.
Our model not only utilizes entities and their latent types as features effectively but also is more interpretable by visualizing attention mechanisms applied to our model and results of LET. 
Experimental results on the SemEval-2010 Task 8, one of the most popular relation classification task, demonstrate that our model outperforms existing state-of-the-art models without any high-level features.
\end{abstract}

\section{Introduction}
Classifying semantic relations between entity pairs in sentences plays a vital role in various NLP tasks, such as information extraction, question answering and knowledge base population \cite{nguyen2015relation}. 
A task of relation classification is defined as predicting a semantic relationship between two tagged entities in a sentence.
\begin{figure}[!ht]
\includegraphics[width=\linewidth]{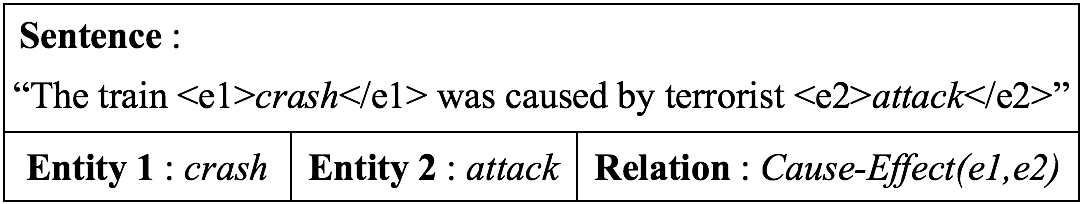}
\caption{A Sample of Relation Classification.}
\label{fig:sample}
\end{figure}\\
For example, given a sentence with tagged entity pair, \textit{crash} and \textit{attack}, this sentence is classified into the relation \textit{Cause-Effect(e1,e2)}\footnote[1]{It is one of the pre-defined relation classes in the SemEval-2010 Task 8 \cite{hendrickx2009semeval}.} between the entity pair like Figure~\ref{fig:sample}.
A first entity is surrounded by $\langle e1\rangle$ and $\langle /e1\rangle$, and a second entity is surrounded by $\langle e2\rangle$ and $\langle /e2\rangle$.

Most previous relation classification models rely heavily on high-level lexical and syntactic features obtained from NLP tools such as WordNet, dependency parser, part-of-speech (POS) tagger, and named entity recognizer (NER).
The classification models relying on such features suffer from propagation of implicit error of the tools and they are computationally expensive.

Recently, many studies therefore propose end-to-end neural models without the high-level features.
Among them, attention-based models, which focus to the most important semantic information in a sentence, show state-of-the-art results in a lot of NLP tasks.
Since these models are mainly proposed for solving translation and language modeling tasks, they could not fully utilize the information of tagged entities in relation classification task.
However, tagged entity pairs could be powerful hints for solving relation classification task.
For example, even if we do not consider other words except the \textit{crash} and \textit{attack}, we intuitively know that the entity pair has a relation \textit{Cause-Effect(e1,e2)}\footnotemark[1] better than \textit{Component-Whole(e1,e2)}\footnotemark[1] in Figure~\ref{fig:sample}

To address these issues, We propose a novel end-to-end recurrent neural model which incorporates an entity-aware attention mechanism with a latent entity typing (LET).
To capture the context of sentences, We obtain word representations by self attention mechanisms and build the recurrent neural architecture with Bidirectional Long Short-Term Memory (LSTM) networks.
Entity-aware attention focuses on the most important semantic information considering entity pairs with word positions relative to these pairs and latent types obtained by LET.

\begin{figure*}[!ht]
\includegraphics[width=\textwidth, height=8.5cm]{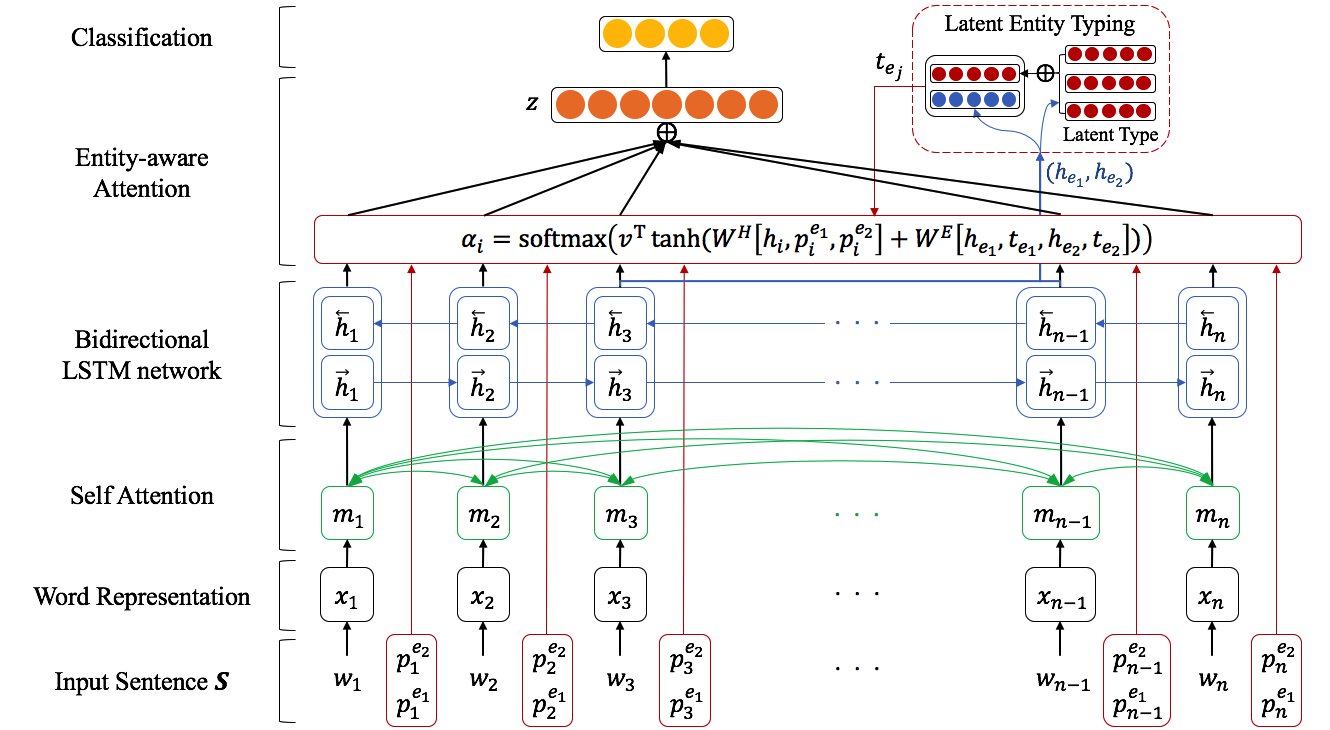}
\caption{The architecture of our model (best viewed in color). Entity 1 and 2 corresponds to the $3$ and $(n-1)$-th words, respectively, which are fed into the LET.}
\label{fig:model}
\end{figure*}

The contributions of our work are summarized as follows:
(1) We propose an novel end-to-end recurrent neural model and an entity-aware attention mechanism with a LET which focuses to semantic information of entities and their latent types;
(2) Our model obtains 85.2\% F1-score in SemEval-2010 Task 8 and it outperforms existing state-of-the-art models without any high-level features;
(3) We show that our model is more interpretable since it's decision making process could be visualized with self attention, entity-aware attention, and LET.

\section{Related Work}
There are several studies for solving relation classification task.
Early methods used handcrafted features through a series of NLP tools or manually designing kernels \cite{rink2010utd}.
These approaches use high-level lexical and syntactic features obtained from NLP tools and manually designing kernels, but the classification models relying on such features suffer from propagation of implicit error of the tools.

On the other hands, deep neural networks have shown outperform previous models using handcraft features. 
Especially, many researches tried to solve the problem based on end-to-end models using only raw sentences and pre-trained word representations learned by Skip-gram and Continuous Bag-of-Words \cite{mikolov2013distributed, mikolov2013efficient, pennington2014glove}.
Zeng et al. employed a deep convolutional neural network (CNN) for extracting lexical and sentence level features \cite{zeng2014relation}.
Dos Santos et al. proposed model for learning vector of each relation class using ranking loss to reduce the impact of artificial classes \cite{dos2015classifying}.
Zhang and Wang used bidirectional recurrent neural network (RNN) to learn long-term dependency between entity pairs \cite{zhang2015relation}.
Furthermore, Zhang et al. proposed bidirectional LSTM network (BLSTM) utilizing position of words, POS tags, named entity information, dependency parse \cite{zhang2015bidirectional}.
This model resolved vanishing gradient problem appeared in RNNs by using BLSTM. 

Recently, some researcher have proposed attention-based models which can focus to the most important semantic information in a sentence.
Zhou et al. combined attention mechanisms with BLSTM \cite{zhou2016attention}.
Xiao and Liu split the sentence into two entities and used two attention-based BLSTM hierarchically \cite{xiao2016semantic}.
Shen and Huang proposed attention-based CNN using word level attention mechanism that is able to better determine which parts of the sentence are more influential \cite{huang2016attention}.

In contrast with end-to-end model, several works proposed models utilizing the shortest dependency path (SDP) between entity pairs of dependency parse trees.
SDP-LSTM model proposed by Yan et al. and deep recurrent neural networks (DRNNs) model proposed by Xu et al eliminate irrelevant words out of SDP and use neural network based on the meaningful words composing SDP \cite{xu2015classifying, xu2016improved}.

\section{Model}
In this section, we introduce a novel recurrent neural model that incorporate an entity-aware attention mechanism with a LET method in detail.
As shown in Figure~\ref{fig:model}, our model consists of four main components: 
(1) \textbf{Word Representation} that maps each word in a sentence into vector representations; 
(2) \textbf{Self Attention} that captures the meaning of the correlation between words based on multi-head attention \cite{vaswani2017attention}; 
(3) \textbf{BLSTM} which sequentially encodes the representations of self attention layer;
(4) \textbf{Entity-aware Attention} that calculates attention weights with respect to the entity pairs, word positions relative to these pairs, and their latent types obtained by LET. 
After that, the features are averaged along the time steps to produce the sentence-level features.

\subsection{Word Representation}
Let a input sentence is denoted by $S=\{w_1, w_2, ..., w_n\}$, where $n$ is the number of words.
We transform each word into vector representations by looking up word embedding matrix $W_{word}\in \mathbb{R}^{d_w\times |V|}$, where $d_w$ is the dimension of the vector and $|V|$ is the size of vocabulary. 
Then the word representations $X=\{x_1, x_2, ..., x_n\}$ are obtained by mapping $w_i$, the $i$-th word, to a column vector $x_i\in \mathbb{R}^{d_w}$ are fed into the next layer.

\subsection{Self Attention}
The word representations are fixed for each word, even though meanings of words vary depending on the context.
Many neural models encoding sequence of words may expect to learn implicitly of the contextual meaning, but they may not learn well because of the long-term dependency problems \cite{bengio1994learning}.
In order for the representation vectors to capture the meaning of words considering the context, we employ the self attention, a special case of attention mechanism, that only requires a single sequence.
Self attention has been successfully applied to various NLP tasks such as machine translation, language understanding, and semantic role labeling \cite{vaswani2017attention, shen2017disan, tan2017deep}.

\begin{figure}[!ht]
\centering
\includegraphics[width=0.98\linewidth]{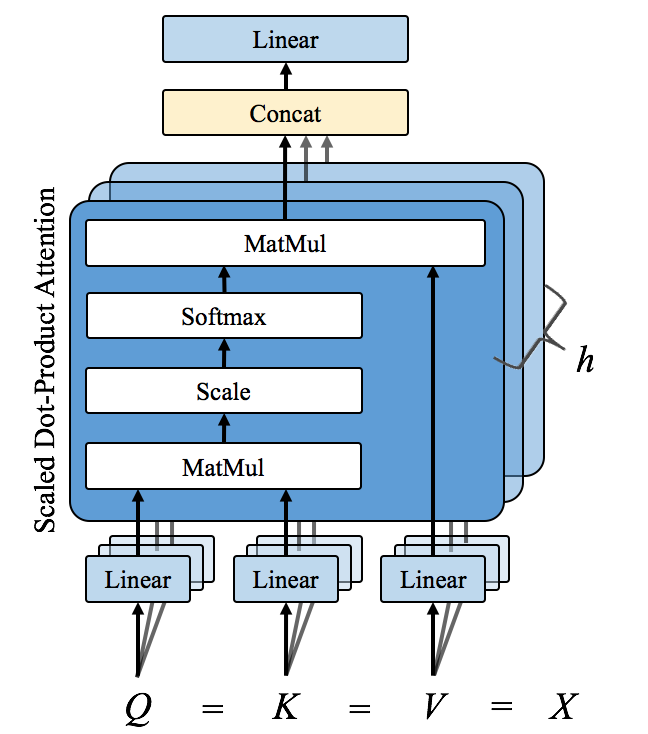}
\caption{Multi-Head Self Attention. For self attention, the $Q$(query), $K$(key), and $V$(value), inputs of multi-head attention, should be the same vectors. In our work, they are equivalent to $X$, the word representation vectors.}
\label{fig:mh}
\end{figure}

We adopt the multi-head attention formulation \cite{vaswani2017attention}, one of the methods for implementing self attentions.
Figure~\ref{fig:mh} illustrates the multi-head attention mechanism that consists of several linear transformations and scaled dot-product attention corresponding to the center block of the figure. 
Given a matrix of $n$ vectors, query $Q$, key $K$, and value $V$, the scaled dot-product attention is calculated by the following equation:
\begin{equation}
\text{Attention}(Q, K, V)=\text{softmax}(\frac{QK^{\top}}{\sqrt{d_w}})V
\label{e3.1}
\end{equation}\\
In multi-head attention, the scaled dot-product attention with linear transformations is performed on $r$ parallel heads to pay attention to different parts.
Then formulation of multi-head attention is defined by the follows:
\begin{equation} 
\text{MultiHead}(Q, K, V)=W^{M}[head_1; ...; head_{r}]
\label{e3.2}
\end{equation}
\begin{equation} 
head_i=\text{Attention}(W_i^QQ, W_i^KK, W_i^VV)
\label{e3.3}
\end{equation}
where [;] indicates row concatenation and $r$ is the number of heads.
The weights $W^{M}\in \mathbb{R}^{d_w\times d_w}$, $W_i^Q\in \mathbb{R}^{d_w/r\times d_w}$, $W_i^K\in \mathbb{R}^{d_w/r\times d_w}$, and $W_i^V\in \mathbb{R}^{d_w/r\times d_w}$ are learnable parameter for linear transformation. $W^{M}$ is for concatenation outputs of scaled dot-product attention and the others are for query, key, value of $i$-th head respectively.

Because our work requires self attention, the input matrices of multi-head attention, $Q$, $K$, and $V$ are all equivalent to $X$, the word representation vectors. As a result, outputs of multi-head attention are denoted by $M=\{m_1, m_2, ..., m_n\}=\text{MultiHead}(X, X, X)$, where $m_i$ is the output vector corresponding to $i$-th word.
The output of self attention layer is the sequence of representations whose include informative factors in the input sentence.

\subsection{Bidirectional LSTM Network}
For sequentially encoding the output of self attention layer, we use a BLSTM \cite{graves2005framewise, graves2013speech} that consists of two sub LSTM networks: a forward LSTM network which encodes the context of a input sentence and a backward LSTM network which encodes that one of the reverse sentence.
More formally, BLSTM works as follows:
\begin{equation}
\overrightarrow{h_t}=\overrightarrow{LSTM}(m_{t})
\label{e3.4}
\end{equation}
\begin{equation}
\overleftarrow{h_t}=\overleftarrow{LSTM}(m_{t})
\label{e3.5}
\end{equation}
\begin{equation} 
h_t=[\overrightarrow{h_t}; \overleftarrow{h_t}]
\label{e3.6}
\end{equation}
The representation vectors $M$ obtained from self attention layer are forwarded into to the network step by step. At the time step $t$, the hidden state $h_t\in \mathbb{R}^{2d_h}$ of a BLSTM is obtained by concatenating $\overrightarrow{h_t}\in \mathbb{R}^{d_h}$, the hidden state of forward LSTM network, and $\overleftarrow{h_t}\in \mathbb{R}^{d_h}$, the backward one, where $d_h$ is dimension of each LSTM's state.

\begin{equation}
    \overrightarrow{h_t}\in \mathbb{R}^{d_h} \: \overleftarrow{h_t}\in \mathbb{R}^{d_h}
\end{equation}
\subsection{Entity-aware Attention Mechanism}
Although many models with attention mechanism achieved state-of-the-art performance in many NLP tasks.
However, for the relation classification task, these models lack of prior knowledge for given entity pairs, which could be powerful hints for solving the task.
Relation classification differs from sentence classification in that information about entities is given along with sentences.

We propose a novel entity-aware attention mechanism for fully utilizing informative factors in given entity pairs. Entity-aware attention utilizes the two additional features except $H=\{h_1, h_2, ..., h_n\}$, (1) relative position features, (2) entity features with LET, and the final sentence representation $z$, result of the attention, is computed as follows:
\begin{equation} 
u_i=\text{tanh}(W^H[h_i;p_{i}^{e_1};p_{i}^{e_2}]+W^E[h_{e_1};t_1;h_{e_2};t_2])
\label{e3.7}
\end{equation}
\begin{equation} 
\alpha_i=\frac{\text{exp}(v^{\top}u_i)}{\sum_{j=1}^{n}\text{exp}(v^{\top}u_j)}
\label{e3.8}
\end{equation}
\begin{equation} 
z=\sum_{i=1}^{n} \alpha_i h_i
\label{e3.9}
\end{equation}

\subsubsection{Relative Position Features}
In relation classification, the position of each word relative to entities has been widely used for word representations \cite{zeng2014relation, nguyen2015relation, huang2016attention}.
Recently, position-aware attention is published as a way to use the relative position features more effectively \cite{zhang2017position}.
It is a variant of attention mechanisms, which use not only outputs of BLSTM but also the relative position features when calculating attention weights.

We adopt this method with slightly modification as shown in Equation~\ref{e3.7}.
In the equation, $p_i^{e_1}\in \mathbb{R}^{d_p}$ and $p_i^{e_2}\in \mathbb{R}^{d_p}$ corresponds to the position of the $i$-th word relative to the first entity ($e_1$-th word) and second entity ($e_2$-th word) in a sentence respectively, where $e_{j\in\{1,2\}}$ is a index of $j$-th entity.
Similar to word embeddings, the relative positions are converted to vector representations by looking up learnable embedding matrix $W_{pos}\in \mathbb{R}^{d_p\times (2L-1)}$, where $d_p$ is the dimension of the relative position vectors and $L$ is the maximum sentence length.

Finally, the representations of BLSTM layer take into account the context and the positional relationship with entities by concatenating $h_i$, $p_i^{e_1}$, and $p_i^{e_2}$.
The representation is linearly transformed by $W^H\in \mathbb{R}^{d_a\times (2d_h+2d_p)}$ as in the Equation~\ref{e3.7}.

\subsubsection{Entity Features with Latent Type}
Since entity pairs are powerful hints for solving relation classification task, we involve the entity pairs and their types in the attention mechanism to effectively train relations between entity pairs and other words in a sentence.
We employ the two entity-aware features. 
The first is the hidden states of BLSTM corresponding to positions of entity pairs, which are high-level features representing entities.
These are denoted by $h_{e_i}\in \mathbb{R}^{2d_h}$, where $e_i$ is index of $i$-th entity.

In addition, latent types of the entities obtained by LET, our proposed novel method, are the second one.
Using types as features can be a great way to improve performance, since the types of entities alone can be inferred the approximate relations.
Because the annotated types are not given, we use the latent type representations by applying the LET inspired by latent topic clustering, a method for predicting latent topic of texts in question answering task \cite{yoon2018latent}.
The LET constructs the type representations by weighting $K$ latent type vectors based on attention mechanisms.
The mathematical formulation is the follows:
\begin{equation} 
a_i^{j}=\frac{\text{exp}((h_{e_j})^{\top}c_i)}{\sum_{k=1}^{K}\text{exp}((h_{e_j})^{\top}c_k)}
\label{e3.10}
\end{equation}
\begin{equation} 
t_{j\in\{1,2\}}=\sum_{i=1}^{K} a_i^{j} c_i
\label{e3.11}
\end{equation}
where $c_i$ is the $i$-th latent type vector and $K$ is the number of latent entity types.

As a result, entity features are constructed by concatenating the hidden states corresponding entity positions and types of entity pairs. After linear transformation of the entity features, they add up with the representations of BLSTM layer as in Equation~\ref{e3.7}, and the representation of sentence $z\in \mathbb{R}^{2d_h}$ is computed by Equations from \ref{e3.7} to \ref{e3.9}.

\subsection{Classification and Training}
The sentence representation obtained from the entity-aware attention $z$ is fed into a fully connected softmax layer for classification. It produces the conditional probability $p(y|S,\theta)$ over all relation types:
\begin{equation} 
p(y|S,\theta)=\text{softmax}(W^Oz+b^O)
\label{e3.12}
\end{equation}
where $y$ is a target relation class and $S$ is the input sentence. The $\theta$ is whole learnable parameters in the whole network including $W^O\in \mathbb{R}^{|R|\times 2d_h}$, $b^O\in \mathbb{R}^{|R|}$, where $|R|$ is the number of relation classes.
A loss function $\mathcal{L}$ is the cross entropy between the predictions and the ground truths, which is defined as:
\begin{equation} 
\mathcal{L}=-\sum_{i=1}^{|D|}\log p(y^{(i)}|S^{(i)},\theta) + \lambda||\theta||_2^2
\label{e3.13}
\end{equation}
where $|D|$ is the size of training dataset and ($S^{(i)}$, $y^{(i)}$) is the $i$-th sample in the dataset. We minimize the loss $\mathcal{L}$ using AdaDelta optimizer \cite{zeiler2012adadelta} to compute the parameters $\theta$ of our model.

To alleviate overfitting, we constrain the L2 regularization with the coefficient $\lambda$ \cite{ng2004feature}. 
In addition, the dropout method is applied after word embedding, LSTM network, and entity-aware attention to prevent co-adaptation of hidden units by randomly omitting feature detectors \cite{hinton2012improving, zaremba2014recurrent}.

\section{Experiments}

\subsection{Dataset and Evaluation Metrics}
We evaluate our model on the SemEval-2010 Task 8 dataset, which is an commonly used benchmark for relation classification \cite{hendrickx2009semeval} and compare the results with the state-of-the-art models in this area.
The dataset contains 10 distinguished relations, \textit{Cause-Effect}, \textit{Instrument-Agency}, \textit{Product-Producer}, \textit{Content-Container}, \textit{Entity-Origin}, \textit{Entity-Destination}, \textit{Component-Whole}, \textit{Member-Collection}, \textit{Message-Topic}, and \textit{Other}.
The former 9 relations have two directions, whereas \textit{Other} is not directional, so the total number of relations is 19.
There are 10,717 annotated sentences which consist of 8,000 samples for training and 2,717 samples for testing.
We adopt the official evaluation metric of SemEval-2010 Task 8, which is based on the macro-averaged F1-score (excluding \textit{Other}), and takes into consideration the directionality.

\subsection{Implementation Details}
We tune the hyperparameters for our model on the development set randomly sampled 800 sentences for validation.
The best hyperparameters in our proposed model are shown in following Table~\ref{tb:1}.
\begin{table}[!ht]
\begin{center}
\renewcommand{\arraystretch}{1.1}
\begin{tabular}{>{\centering\arraybackslash}m{1.8cm} l c}
Hyper-parameter & Description & Value \\ 
\hline\hline
$d_w$ & Size of Word Embeddings & 300 \\ \hline
$r$ & Number of Heads & 4 \\ \hline
$d_h$ & Size of Hidden Layer & 300 \\ \hline
$d_p$ & Size of Position Embeddings & 50 \\ \hline
$d_a$ & Size of Attention Layer & 50 \\ \hline
$K$ & Number of Latent Entity Types & 3 \\ \hline
$batch\_size$ & Size of Mini-Batch & 20 \\ \hline
$\eta$ & Initial Learning Rate & 1.0 \\ \hline
\multirow{3}{1.8cm}{\textit{dropout rate}\centering}
& Word Embedding layer & 0.3 \\
& BLSTM layer& 0.3 \\
& Entity-aware Attention layer & 0.5 \\ \hline
$\lambda$ & L2 Regularization Coefficient & $10^{-5}$ \\ \hline
\end{tabular}
\end{center}
\caption{Hyperparameters.}
\label{tb:1}
\end{table}

We use pre-trained weights of the publicly available GloVe model \cite{pennington2014glove} to initialize word embeddings in our model, and other weights are randomly initialized from zero-mean Gaussian distribution \cite{glorot2010understanding}.

\subsection{Experimental Results}
Table~\ref{tb:2} compares our Entity-aware Attention LSTM model with state-of-the-art models on this relation classification dataset. 
We divide the models into three groups, \textit{Non-Neural Model}, \textit{SDP-based Model}, and \textit{End-to-End Model}.
First, the SVM \cite{rink2010utd}, \textit{Non-Neural Model}, was top of the SemEval-2010 task, during the official competition period. 
They used many handcraft feature and SVM classifier. 
As a result, they achieved an F1-score of 82.2\%.
The second is \textit{SDP-based Model} such as MVRNN \cite{socher2012semantic}, FCM \cite{yu2014factor}, DepNN \cite{liu2015dependency}, depLCNN+NS \cite{xu2015semantic}, SDP-LSTM \cite{xu2015classifying}, and DRNNs \cite{xu2016improved}.
The SDP is reasonable features for detecting semantic structure of sentences.
Actually, the SDP-based models show high performance, but SDP may not always be accurate and the parsing time is exponentially increased by long sentences.
The last model is \textit{End-to-End Model} automatically learned internal representations can occur between the original inputs and the final outputs in deep learning.
There are CNN-based models such as CNN \cite{zeng2014relation, nguyen2015relation}, CR-CNN \cite{dos2015classifying}, and Attention-CNN \cite{huang2016attention} and RNN-based models such as BLSTM \cite{zhang2015bidirectional}, Attention-BLSTM \cite{zhou2016attention}, and Hierarchical-BLSTM (Hier-BLSTM) \cite{yang2016hierarchical} for this task.

\begin{table}[!ht]
\begin{center}
\renewcommand{\arraystretch}{1.1}   
\begin{tabular}{m{1.8cm} p{4.7cm} c}
& \textbf{Model} & \textbf{F1} \\ 
\hline\hline
\textit{Non-Neural Model}\centering
& SVM & 82.2 \\
\hline
\multirow{6}{1.8cm}{\textit{SDP-based Model}\centering} 
& MVRNN & 82.4 \\ \cline{2-3}
& FCM & 83.0 \\ \cline{2-3}
& DepNN & 83.6 \\ \cline{2-3}
& depLCNN+NS & \textbf{85.6} \\ \cline{2-3}
& SDP-LSTM & 83.7 \\ \cline{2-3}
& DRNNs & \textbf{86.1} \\ 
\hline
\multirow{9}{1.8cm}{\textit{End-to-End Model}\centering}
& CNN & 82.7 \\ \cline{2-3}
& CR-CNN & 84.1 \\ \cline{2-3}
& Attention-CNN & 84.3 \\
& + POS, WN, WAN & \textbf{85.9} \\ \cline{2-3}
& BLSTM & 82.7 \\
& + PF, POS, NER, DEP, WN & 84.3 \\ \cline{2-3}
& Attention-BLSTM & 84.0 \\ \cline{2-3}
& Hier-BLSTM & 84.3 \\ 
\hline
& \textbf{Our Model} & \textbf{84.7} \\
& \textbf{+ Latent Entity Typing} & \textbf{85.2} \\ \hline
\end{tabular}
\end{center}
\caption{Comparison with Previous Results on SemEval-2010 Task 8 dataset, where the WN, WAN, PF, and DEP are WordNet (hypernyms), words around nominals, position features, and dependency features, respectively.}
\label{tb:2}
\end{table}
Our proposed model achieves an F1-score of 85.2\% which outperforms all competing state-of-the-art approaches except depLCNN+NS, DRNNs, and Attention-CNN.
However, they rely on high-level lexical features such as WordNet, dependency parse trees, POS tags, and NER tags from NLP tools.

The experimental results show that the LET is effective for relation classification. 
The LET improve a performance of 0.5\% than the model not applied it. The model showed the best performance with three types.

\section{Visualization}
There are three different visualization to demonstrate that our model is more interpretable.
First, the visualization of self attention shows where each word focus on parts of a sentence.
By showing the words that the entity pair attends, we can find the words that well represent the relation between them.
Next, the entity-aware attention visualization shows where the model pays attend to a sentence.
This visualization result highlights important words in a sentence, which are usually important keywords for classification.
Finally, we visualize representation of type in LET by using t-SNE \cite{maaten2008visualizing}, a method for dimensionality reduction, and group the whole entities in the dataset by the its latent types.


\subsection{Self Attention}
\begin{figure}[!ht]
\includegraphics[width=\linewidth]{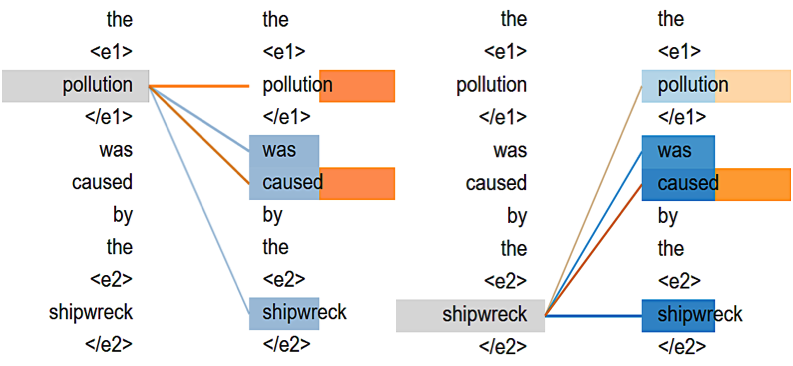}
\caption{Visualization of Self Attention.}
\label{fig:self_att}
\end{figure}

We can obtain the richer word representations by using self attentions.
These word representations are considered the context based on correlation between words in a sentence.
The Figure~\ref{fig:self_att} illustrates the results of the self attention in the sentence, ``\textit{the \textlangle e1\textrangle pollution\textlangle /e1\textrangle was caused by the \textlangle e2\textrangle shipwrek\textlangle /e2\textrangle}'', which is labeled \textit{Cause-Effect(e1,e2)}.
There are visualizations of the two heads in the multi-head attention applied for self attention.
The color density indicates the attention values, results of Equation~\ref{e3.1}, which means how much an entity focuses on each word in a sentence.

In Figure~\ref{fig:self_att}, the left represents the words that \textit{pollution}, the first entity, focuses on and the right represents the words that \textit{shipwreck}, the second entity, focuses on.
We can recognize that the entity pair is commonly concentrated on \textit{was}, \textit{caused}, and each other.
Actually, these words play the most important role in semantically predicting the \textit{Cause-Effect(e1,e2)}, which is the relation class of this entity pair.

\subsection{Entity-aware Attention}
\begin{figure*}[!ht]
\includegraphics[width=\textwidth]{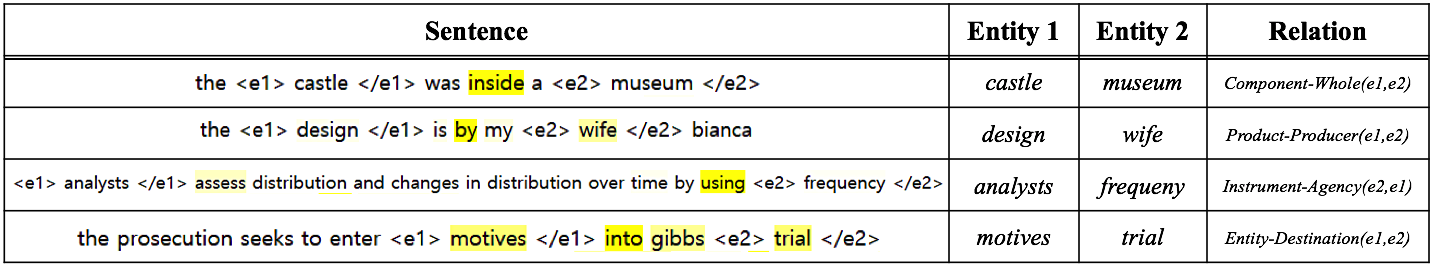}
\caption{Visualization of Entity-aware Attention}
\label{fig:att_viz}
\end{figure*}
Figure~\ref{fig:att_viz} shows where the model focuses on the sentence to compute relations between entity pairs, which is the result of visualizing the alpha vectors in Equation~\ref{e3.8}.
The important words in sentence are highlighted in yellow, which means that the more clearly the color is, the more important it is.
For example, in the first sentence, the \textit{inside} is strongly highlighted, which is actually the best word representing the relation \textit{Component-whole(e1,e2)} between the given entity pair.
As another example, in the third sentence, the highlighted \textit{assess} and \textit{using} represent the relation, \textit{Instrument-Agency(e2,e1)} between entity pair, \textit{analysts} and \textit{frequency}, well.
We can see that the \textit{using} is more highlighted than the \textit{assess}, because the former represents the relation better.

\subsection{Latent Entity Type}
\begin{figure}[!ht]
\includegraphics[width=\linewidth]{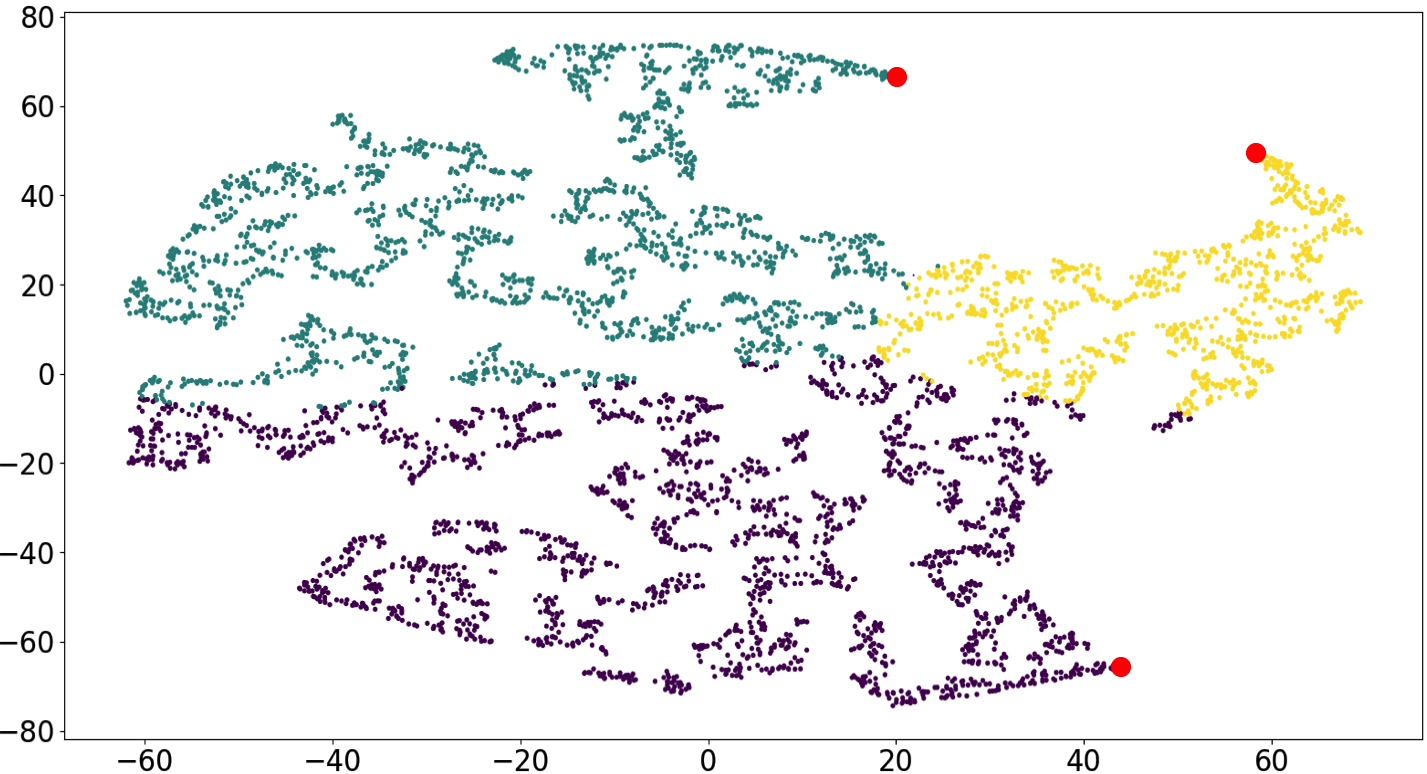}
\caption{Visualization of latent type representations using t-SNE}
\label{fig:type_viz}
\end{figure}
Figure~\ref{fig:type_viz} visualizes latent type representation $t_{j\in \{1,2\}}$ in Equation~\ref{e3.11}
Since the dimensionality of representation vectors are too large to visualize, we applied the t-SNE, one of the most popular dimensionality reduction methods.
In Figure~\ref{fig:type_viz}, the red points represent latent type vectors $c_{i\in K}$ and the rests are latent type representations $t_j$, where the colors of points are determined by the closest of the latent type vectors in the vector space of the original dimensionality.
The points are generally well divided and are almost uniformly distributed without being biased to one side.

\begin{figure}[!ht]
\includegraphics[width=\linewidth]{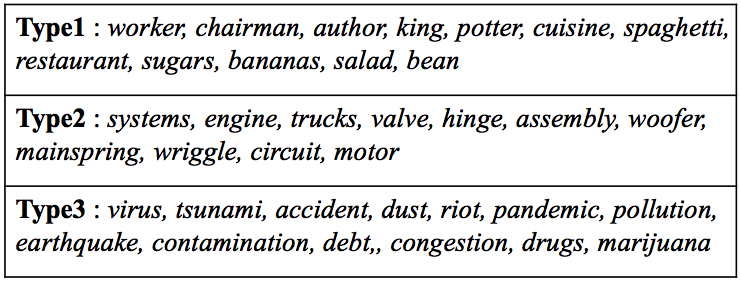}
\caption{Sets of Entities grouped by Latent Types}
\label{fig:type}
\end{figure}
Figure~\ref{fig:type} summarizes the results of extracting 50 entities in close order with each latent type vector.
This allows us to roughly understand what latent types of entities are.
We use a total of three types and find that similar characteristics appear in words grouped by together.
In the type 1, the words are related to human's jobs and foods.
The type2 has a lot of entities related to machines and engineering like \textit{engine}, \textit{woofer}, and \textit{motor}.
Finally, in type3, there are many words with bad meanings related associated with disasters and drugs.
As a result, each type has a set of words with similar characteristics, which can prove that LET works effectively.

\section{Conclusion}
In this paper, we proposed entity-aware attention mechanism with latent entity typing and a novel end-to-end recurrent neural model which incorporates this mechanism for relation classification.
Our model achieves 85.2\% F1-score in SemEval-2010 Task 8 using only raw sentence and word embeddings without any high-level features from NLP tools and it outperforms existing state-of-the-art methods.
In addition, our three visualizations of attention mechanisms applied to the model demonstrate that our model is more interpretable than previous models.
We expect our model to be extended not only the relation classification task but also other tasks that entity plays an important role.
Especially, latent entity typing can be effectively applied to sequence modeling task using entity information without NER.
In the future, we will propose a new method in question answering or knowledge base population based on relations between entities extracted from our model.

\bibliographystyle{siam}
\bibliography{refs.bib}

\end{document}